\documentclass{article}
\usepackage{nips13submit_e,times}
\usepackage[pdftex]{graphicx}
\usepackage[utf8]{inputenc}
\usepackage{amsmath, authblk, subfig}
\usepackage{hyperref}
\usepackage{booktabs}
\usepackage{microtype}
\usepackage{caption}
\usepackage{wrapfig}

\title{An investigation into language complexity of World-of-Warcraft game-external texts}

\author{
\textbf{Simon Šuster}\\
Computational Linguistics\\
University of Groningen\\
\texttt{s.suster@rug.nl}\\
}

                                                                                                                                                                                                                          \nipsfinalcopy 

\begin{document}
\maketitle
\begin{abstract}
We present a language complexity analysis of World of Warcraft (WoW) community texts, which we compare to texts from a general corpus of web English. Results from several complexity types are presented, including lexical diversity, density, readability and syntactic complexity. The language of WoW texts is found to be comparable to the general corpus on some complexity measures, yet more specialized on other measures. Our findings can be used by educators willing to include game-related activities into school curricula.
\end{abstract}
\section{Introduction}
In computer-assisted language learning, there is a relatively substantial body of work relating to the ways and possible usefulness of including video games in the language learning process. Thus, types and characteristics of in-game communication, the social interaction, and dynamics and communication patterns between gamers are well described. However, very little research has been done on the properties of the text production itself in games and of gaming community. In our opinion, the potential of games for language learning has been largely overlooked by educators, and by researching this language more intensively -- both its social aspects and on large collections of texts -- we can get to know more precisely in what ways the gaming language can be useful in language learning.

This paper focuses on a massively multi-player online game (MMOG), namely World of Warcraft (WoW).\footnote{It is one of the most popular online games ever with presently around 8 million subscribers (see \url{bit.ly/NCMe4E}).} This research focuses on describing automated ways for analysis of language production of the WoW community and potential problems associated with these techniques. Naturally, various measures can be adopted to determine language complexity. Here, we use measures of lexical richness, lexical density, readability and syntactic complexity. We include in our analysis only game-external texts, which we compare to general English on the web.



\section{Background: games as a language-learning tool}
MMOGs are game worlds characterized by its online character, meaning that gamers can interact not only with the game software but also with other players. Collaboration takes place and can be observed both outside the game in various discussion fora and fan sites, and in the game itself. Gaming in the MMOGs is cognitively demanding; it involves exploration of complex problem spaces, constructing and analyzing models, and negotiating meaning within the gaming community; it involves coordinating people, virtual tools and artifacts, and it is also characterized by various forms of texts \cite{steinkuehler}. 


Thorne~\cite{Thorne2008} and Thorne et al.~\cite{ThorneBlackSykes} looked at the intercultural communication in massively multi-player online games, especially WoW. They notice that WoW is played in language-specific domains that concentrate together speakers of the same language. The advantage of this is that after installing a language pack, the game setting and most probably gamers as well will be language specific. However, this may have the disadvantage of hindering multilingual communication in the game. Thorne~\cite{Thorne2008} provides a case study in which two gamers with different levels of gaming proficiency exchange expert knowledge, language-specific explicit corrections and requests for help. Involving in games often results in increased motivation for L2 learning, as also noted by Thorne. 

The work most similar to ours is Thorne et al.~\cite{ThorneEtAl2012}, an exploratory study of linguistic quality and complexity of both internal and external texts of the WoW. By analyzing the complexity of texts gamers engage with, the authors aim to ascertain the relevance of MMOGs for L2 development. In comparison to Thorne et al., our work has a narrower focus and differs in that: a) we provide additional details from the computational analysis of gaming texts, i.e. we also investigate lexical density and fit a probabilistic model for vocabulary growth rate, and b) we contrast the complexity of game texts to diverse web texts from a reference corpus of English.   

Rankin et al.~\cite{RankinGoldGooch} claim that MMOGs, especially their role-playing sub-type, are particularly important for language learning because they represent immersive environments allowing fully-fledged experience of a virtual world. They also abound with possibilities for social interaction, they are motivating and they create a virtual world as the context where language students concentrate on accurate and coherent language use, develop and test their practices. Gee~\cite{gee2007good} sees the language use in games as one developing a complex specialist language, by which children develop their ``islands of expertise''. Rankin et al.~\cite{Rankin2008} and Bryant~\cite{BryantWoW} report on successful inclusion of game-related activities in school curricula.

Here, we do not include in the discussion the educational and conversation games.\footnote{Such as the open-world adventure game Bot Colony (\url{www.botcolony.com}) which encourages user participation in conversation, provides corrective feedback etc.} The availability of truly engaging educational games is as of today still limited \cite{WilcoxGames}.

In this section, we have reviewed the literature that deals mostly with the social aspects of language learning in games. In the next section, we move on to discuss the array of techniques used in this paper to describe the text production of the WoW gaming community. 

\section{Measuring language complexity}
We first treat the measures emphasizing lexical aspects of language production, and then present a couple of measures that take into account syntactic complexity.
\subsection{Lexical measures}
\paragraph{Lexical diversity}
Lexical diversity measures the size of vocabulary in a text or of an author. It is often used for describing and predicting vocabulary growth \cite{baayen}. Intuitively, lexical diversity should be quantifiable in terms of different word types. Since this measure is known to depend heavily on text size, number of tokens are also taken into account. A measure using these two types of information is type-token ratio (TTR). However, even TTR is dependent on the text size: it is bigger when texts are small and decreases as the texts get larger. Due to this fact, the comparison is warranted only on same-sized texts or using statistical approximations (corrected ratios) that take this into account, or using probabilistic models. There exist several corrected measures, like Herdan's C, Guiraud's R, Yule's K and others \cite{baayen}. But since the within-text variability can be large (words are not used randomly), these measures may not be precise at differing sizes of text samples, as shown by Tweedie and Baayen~\cite{TweedieBaayen}. Similar applies to probabilistic models, such as Sichel’s generalized inverse Gauss-Poisson model and Zipf-Mandelbrot models \cite{zipfR, TweedieBaayen}. Especially the latter performed reasonably well and proved length-invariant in the Tweedie and Baayen's~\cite{TweedieBaayen} research. These are also called models for Large Number of Rare Events (LNRE) distributions.

A simple transformation of TTR is mean segmental type-token ratio (MSTTR), calculated by first computing TTR on same-sized text samples, e.g. of 50 tokens, and then taking the mean \cite{LuThorne2011}. As the mean here drowns out the individual differences in samples, all sample-TTRs can be plotted to illustrate the ratio values changing as a function of the position in text.

\paragraph{Lexical density}
Lexical density has been shown to correlate strongly with lexical diversity \cite{Johansson}. It measures the proportion of content, or lexical, words (verbs, nouns, etc.) to the total number of tokens, and it is an indicator of information packaging in the text. It is not trivial to operationalize the definition of lexical density. First, one should decide what counts as a lexical word. Usually, lexical words are said to occur in closed word classes, like nouns, but there is some controversy, for example, whether it makes sense to count some adverbs as lexical or non-lexical words. Secondly, an example like ``turn up'' may consist of a lexical word and a non-lexical function word. A slightly more advance system could count phrases as a single lexical item, see Johansson~\cite{Johansson} for further discussion.
\paragraph{Readability}
Readability formulas try to estimate the degree of text difficulty. There are over 200 of such formulas \cite{dubay}. Many of those include a combination of sentence complexity -- usually mean length of the sentence -- and word complexity measures -- usually average number of syllables or characters per word, or the proportion of complex words. Lu et al.~\cite{LuThorne2011} predict appropriate grade levels of school reading materials successfully with Flesch Reading Ease, Coleman-Liau and New Dale-Chall Readability formulas. Flesch Reading Ease, the formula used in this research, is a function over the number of both syllables and sentences. The reading ease is predicted on a scale from 0 to 100, with 30 being very difficult and 70-and-up being easy. The index is calculated by: $\text{Flesch Reading Ease} = 206.835 - 1.015(N_{\text{Words}}/N_{\text{Sentences}}) - 84.6(N_{\text{Syllables}}/N_{\text{Words}})$.
A modified formula to produce a US grade-level score is called Flesch-Kincaid: $\text{Flesch-Kincaid} = 0.39 (N_{\text{Words}}/N_{\text{Sentences}}) + 11.8 (N_{\text{Syllables}}/N_{\text{Words}}) - 15.59$

\subsection{Syntactic-complexity measures}
The simplest measure of syntactic complexity (or rather its proxy) is the mean length of sentences. Although insensitive to structural differences within sentences, it turns out as a reliable indicator of grade level \cite{LuThorne2011}. There exist more complex measures that define the developmental level of the sentences (D-level, DSS scoring, IPSyn) \cite{LuThorne2011}, and descriptive techniques, such as comparing the frequencies of POS tags for the most common n-grams between two texts \cite{Nerbonnesyntactic}\label{syntactic_complexity}, capable of qualitatively accounting for differences in frequency of syntactic constructions in texts. D-level,  used in this paper, originates in the child language acquisition research and follows the premise that the most complex sentence types are acquired last \cite{Covington}. According to Cheung and Kemper~\cite{Cheung}, D-level is a more adequate index of sentence complexity than other metrics since it is able to differentiate between types of clausal embedding.
\section{Experimental setup}
\paragraph{Gathering data} We construct a corpus of texts from WoW fora, wikis and, to a lesser extent, news. The compilation was carried out with the BootCat toolkit \cite{bootcat}. The corpus contains 261,000 tokens from 196 URLs. The second corpus used as benchmark is the ukWaC \cite{ukwac}. This corpus was built from diverse web-based texts by querying the .uk Internet domain. The original size of ukWaC is 1.9 billion tokens, from which we sample 261,000-token worth of sentences. The raw corpora were lemmatized and POS-tagged with TreeTagger \cite{treetagger}.
\paragraph{Processing data} In this research, the following measures were calculated: TTR; MSTTR; N, V and Adj density; Flesch Reading Ease and Flesch-Kincaid indices; Mean sentence length; and D-level (only for the WoW corpus). In addition, we fit a LNRE model to the data, which allows comparison of differently sized texts (although not strictly needed here) by interpolation (for smaller samples) or extrapolation (for larger samples).

For fitting and plotting probability models, the \verb|zipfR| package \cite{zipfR} was used. All the measures listed above, except D-level, were calculated in R using the \verb|koRpus| package \cite{koRpus}. For testing significance of the difference between the two corpora in vocabulary size and vocabulary growth, Z-statistic from \verb|languageR| \cite{languageR} package was used.\footnote{The variance here is estimated internally by fitting an LNRE model to two texts with a $\chi^2$ test.} When comparing different distributions, as in the case of TTR on text samples and Flesch Reading Ease index on samples, the Kolmogorov-Smirnov test was used to determine the significance of separation between two distributions, cf.~Baayen~\cite{baayen}.

D-level scores were calculated with D-Level Analyzer\footnote{It outputs for each sentence a number between 0 and 7, corresponding to developmental levels (see figure \ref{D-levels}.} \cite{Lu}. The analyzer required POS tagging of raw texts using Penn Treebank tagset and also syntactic annotation with the Collins' parser. Due to some incompatibilities between POS-tag scheme and the parser, syntactic analysis was was often erratic, and some issues remained unresolved such as long sentences returning errors when parsed. We therefore included only 2209 sentences (out of 14491) in the analysis.


\section{Results}
\subsection{Lexical diversity}
The table~(\ref{lexical}) shows main results for WoW and ukWaC corpora. The type-token ratio for the entire $\sim$230,000-token\footnote{During the calculation of these scores, the punctuation is removed, so the total token count is smaller.} WoW corpus is .09, while the TTR for the ukWaC corpus is 0.12. In other words, the probability of encountering a new type at the end of the WoW corpus is 9 per cent, and in ukWaC, this is only 3 per cent more likely. 

\begin{wraptable}{r}{5.5cm}
\caption{Lexical diversity and density results.}\label{lexical}
\begin{tabular}{l c c}
& WoW corpus & ukWaC\\
\cmidrule{1-3}
TTR & .09 & .12\\

MSTTR & .73 & .74\\

N ratio & .16 & .16\\

V ratio & .14 & .15\\

Adj ratio & .06 & .08\\
\cmidrule{1-3}
\end{tabular}

\end{wraptable}

However, having just one ratio value for the entire corpus only gives us a rough picture. To reveal the distribution of TTRs on smaller samples, the mean-segmental TTR was also calculated on samples of 100 tokens. Thus, the MSTTR (mean of 2,300 samples) for the WoW corpus is .73, and .74 for ukWaC. Thus, on average, the probability of encountering a new type after a 100-token sample is 73 per cent, and only 1 per cent higher in ukWaC. Since MSTTR is a mean summary of TTRs of samples, it is useful to show the dispersion of the values graphically by estimating the density curve (cf. \cite{baayen}). 

\begin{figure}[h!]
\begin{center}
\caption{TTR for ukWaC and WoW 100-token samples}\label{msttr}
\includegraphics[width=0.6\textwidth]{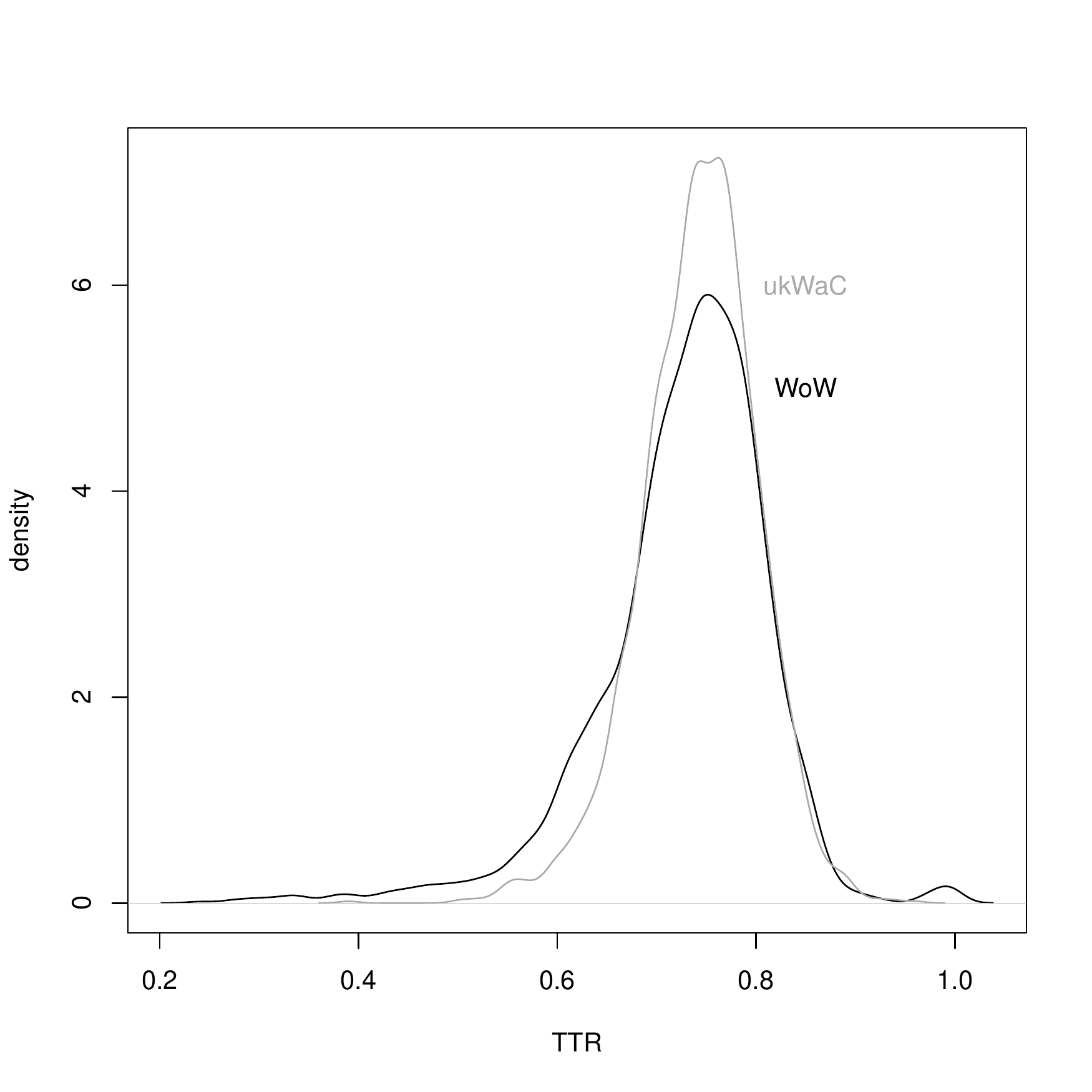}
\end{center}

\end{figure}

\begin{figure}[h!]
\begin{center}
\caption{Flesch Reading Ease index for the WoW and ukWaC corpora}\label{flesch}
\includegraphics[width=0.6\textwidth]{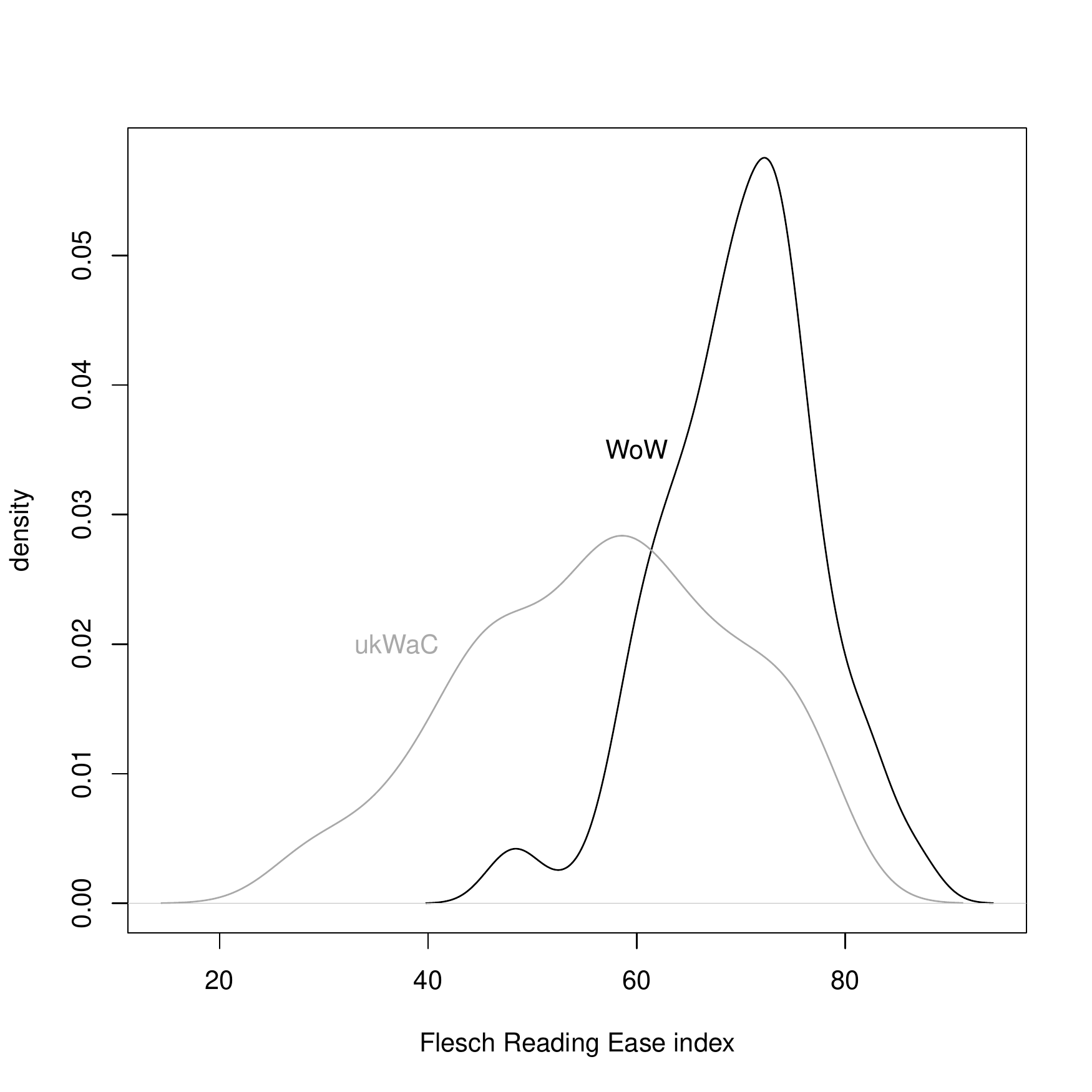}
\end{center}
\end{figure}

Graph~(\ref{msttr}) contains two curves showing the distribution of TTR values on small samples. It is easy to see that there is greater dispersion of TTRs in the case of WoW, whereas the ukWaC values are more uniform, i.e. the curve on the graph has a higher and narrower peak. This difference is summarized by standard deviation, $SD_{WoW}=.087$; $SD_{ukWaC}=.059$. We can attribute these results to greater vocabulary variation in WoW corpus texts, i.e. more texts with lower vocabulary size, but they can also be explained by peculiarities of sampling web texts for WoW corpus.\footnote{E.g. repetitions in internet fora (referring to others' posts) leading to more samples having lower TTR.} The difference between the distributions was statistically significant $p<.001$ (Kolmogorov-Smirnoff $D=.1$). 

\subsection{Probabilistic model}
Sometimes, it may also be interesting or necessary to look at the vocabulary growth rate, because texts may include the same number of types, but with a different rate of occurrence of unseen types \cite{baayen}. More broadly, introduction of new types, or counting the hapax legomena in a sample, is indicative of how productive the language or particular facets of language are. 

\begin{figure}
\centering
\caption{Vocabulary growth for the two corpora. The highest curve on the plot represents the total number of types (observed and expected), the lower curve represents the number of types occurring only once (hapax legomena), even lower curve the number of dis-legomena, and so on.}\label{growth}
\subfloat[WoW]{\includegraphics[width=0.6\textwidth]{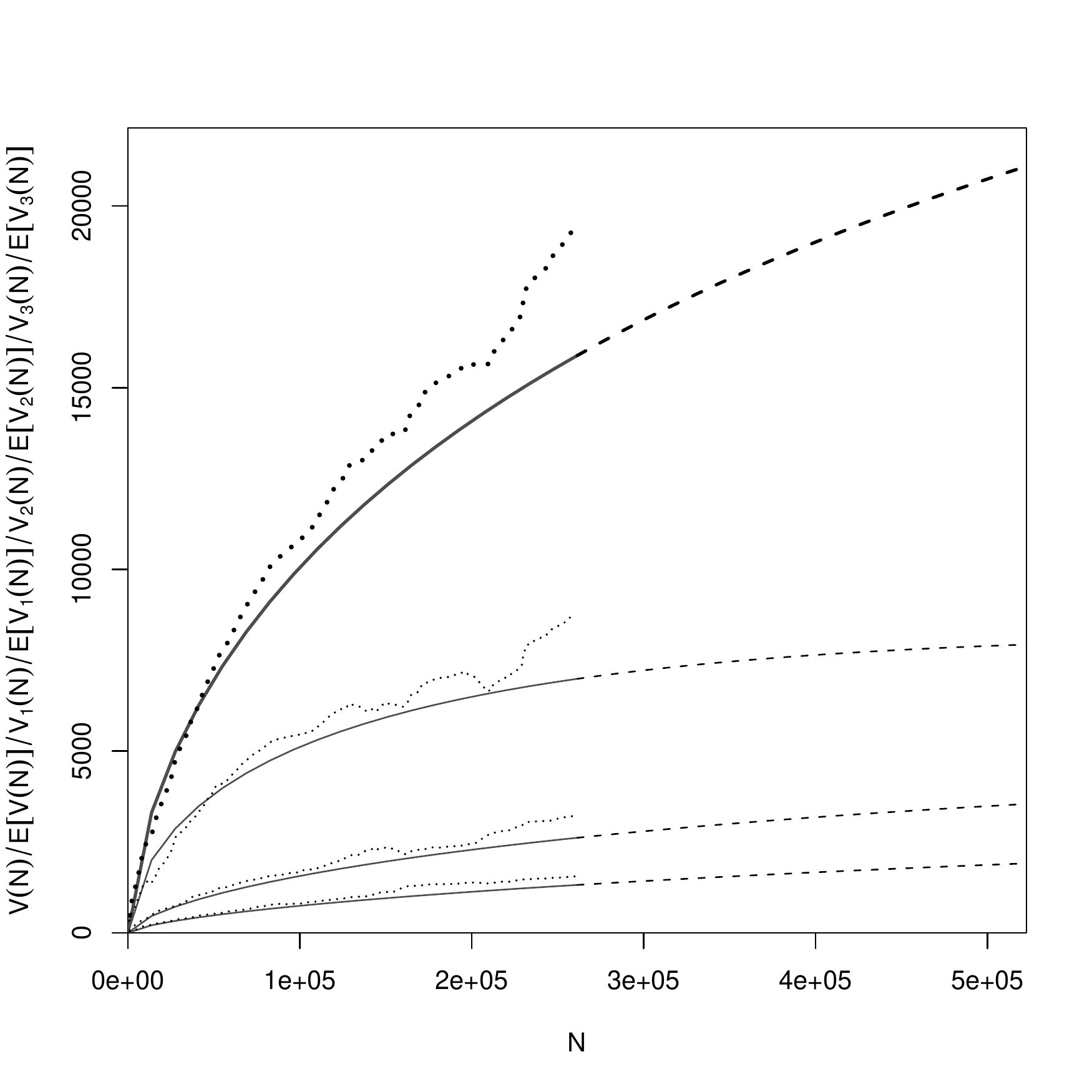}}
\hspace{5mm}
\subfloat[ukWaC]{\includegraphics[width=0.6\textwidth]{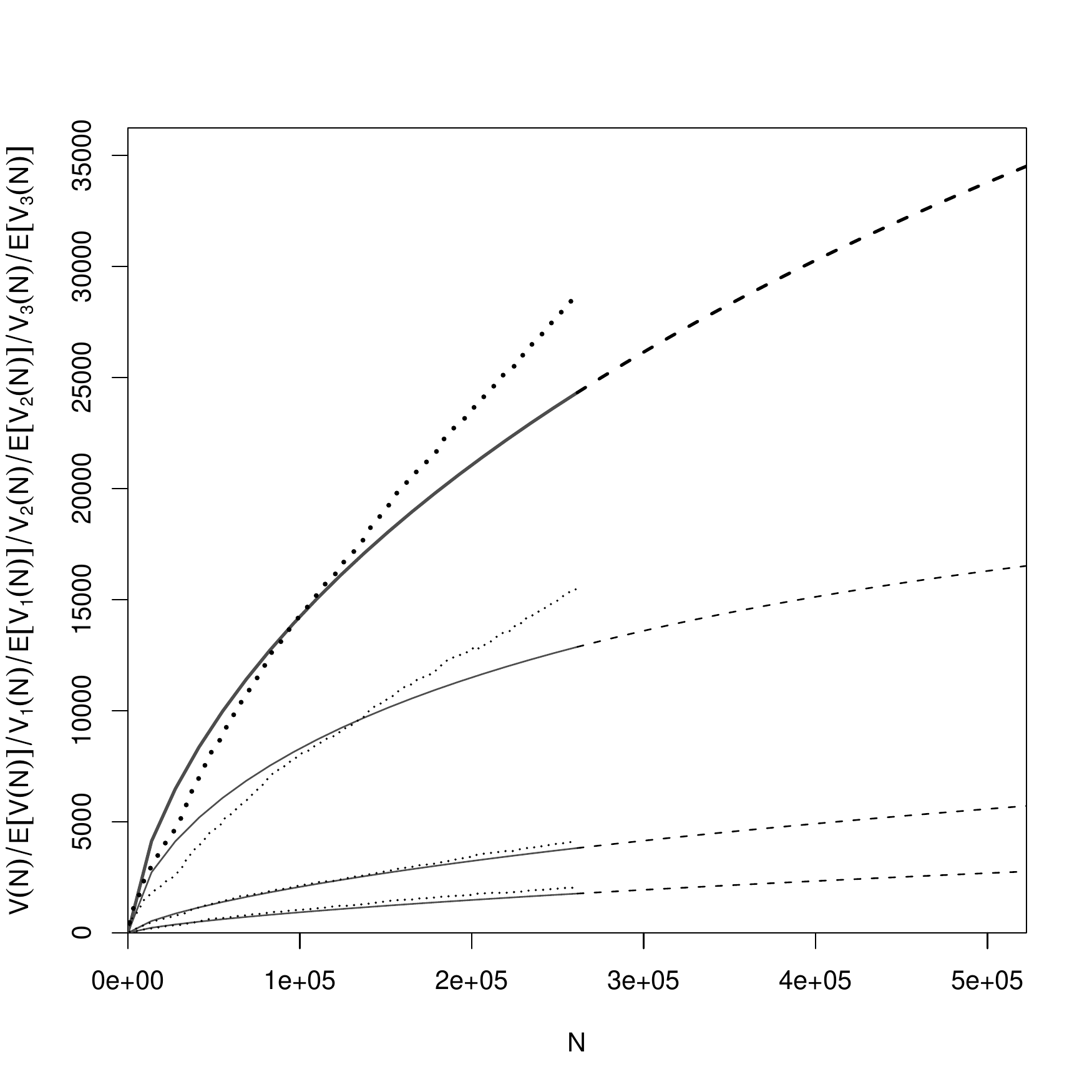}}
\end{figure}

In our case, the vocabulary growth rate is estimated by the ratio of the number of hapax legomena to the number of tokens.  The results of the growth test for the two corpora are $growth.rate_{WoW}=.033, growth.rate_{ukWaC}=.059$. The difference between the results was determined with the Z-statistic, and was significant with $p$ being effectively 0 ($Z=-50.2$). Since the size of the compared corpora is the same, the growth rate essentially tells us that there are much less (almost by half) hapaxes in the WoW corpus than in the ukWaC. This finding can be explained by the nature of the corpora: ukWaC is a general, broad-coverage corpus, while WoW is a specialized, one-domain corpus, so a smaller number of new types is introduced. Graphs \ref{growth} show the vocabulary growth, i.e. the number of new types encountered when moving linearly through the corpus. The dotted line shows observed counts, and the smooth line the expected values, where its solid part indicates interpolated values and the dashed part extrapolated values. For both corpora, we can see that the expected curve is not very accurate. In these graphs, we fitted the inverse generalized Gauss-Poisson model to the data, which yielded a closer fit than the Zipf-Mandelbrot models, but the result is still not satisfactory: observed curves are mostly high above the curve of expected values; also, new types are sampled more slowly than expected if words were used randomly: the observed curve is under the expected one for around 40,000 tokens in WoW and 100,000 tokens in ukWaC, but following these tokens, the expected growth is clearly underestimated. These two observations may be due to the heavy use of topical words when the discourse is cohesive (words are not randomly chosen in a cohesive discourse), but new texts are constantly being introduced when we move from the beginning to the end of the corpus, which may explain the consistent introduction of more and more new types (note how the distance between observed and expected increases). 

Another noticeable fact is that the observed curve for WoW is less smooth than for ukWaC. Such sharp shifts in the curve are unlikely to occur in ``normally'' structured cohesive texts. A possible reason is an inaccurate sampling procedure. From what we have noted in this section, it would be impossible to accurately predict the number of hapax (new types), dis-, etc. legomena for larger or smaller samples based on extrapolated or interpolated values.

\subsection{Lexical density}
Lexical density was computed separately for nouns, verbs and adjectives (table \ref{lexical}), but cumulatively for all texts in each corpus. The proportion of nouns (excluding proper nouns) is .16 in both corpora. There are fewer verbs, .14 and .15 out of all tokens for WoW and ukWaC, respectively. The difference is bigger for the proportion of adjectives, .06 and .08. These data suggest that there is the same amount of words carrying ``nominal'' information, whereas the WoW corpus contains smaller proportions of verbs and adjectives. This can be interpreted as the communication in WoW texts being closer to spoken, and ukWaC texts closer to written language, although the difference is minimal. The observation that spoken texts tend to have lower lexical density was made already by Halliday~\cite{halliday}.

\subsection{Readability}
The Flesch Reading Ease index was calculated on 1000-token samples. From graph~(\ref{flesch}), it can be seen that the results look considerably different. The values for WoW texts are more uniform, while the ukWaC values show greater dispersion, $SD_{WoW}=7.6$; $SD_{ukWaC}=13$. This is not surprising, since ukWaC contains more diverse and genre-rich texts with varying degrees of readability. Compared to ukWaC, WoW corpus is then a homogeneous mass with most texts having similar readability. The difference between the means of Flesch scores for two corpora is 13 points, $M_{WoW}=70$; $M_{ukWaC}=57$. This difference is significant at $p<.001$ ($D=.52$). By Flesch-Kincaid grading system, this finding means that WoW texts are 2.2-grade levels below the ukWaC corpus, $M_{WoW}=7.8$; $M_{ukWaC}=10$. Flesch's classification (table \ref{flesch_grades} in Appendix) of readability of adult reading materials puts these results into a wider perspective. Thus, WoW texts can be described as ``standard-style'', understood on average by $\sim$83\% of US adults.\footnote{Of course, this is only a very rough estimate due to datedness of the Flesch's research on readability.} ukWaC texts are characterized as ``fairly difficult'', presumably understood on average by 54\% of US adults. The Flesch-Kincaid score of 7.8  for WoW sets its texts at the reading level of the age group 12--14 and higher.\footnote{In other words, for students of US middle schools (or European secondary schools).} 

\subsection{Sentence length}
\begin{wraptable}{r}{5.5cm}
\caption{Mean sentence length for the WoW and ukWaC corpora.}\label{msl}
\begin{tabular}{l c c}
& WoW corpus & ukWaC \\
\cmidrule{1-3}
MSL (SD) & 19 (14) & 21 (14)\\
\end{tabular}
\end{wraptable}
The Flesch index, as mentioned before, already incorporates one syntactic information, i.e. the mean sentence length. There are on average 19 words per sentence in WoW texts (see table~\ref{msl}), and two words more in an average ukWaC sentence. According to the table (\ref{flesch_grades}), the score for the WoW texts corresponds to style categories ``standard'' and ``fairly difficult''.

\begin{figure}
\caption{D-level scores expressed in number of sentences that fall into a particular level.}\label{d-level-wow}
\begin{center}
\includegraphics[width=0.5\textwidth]{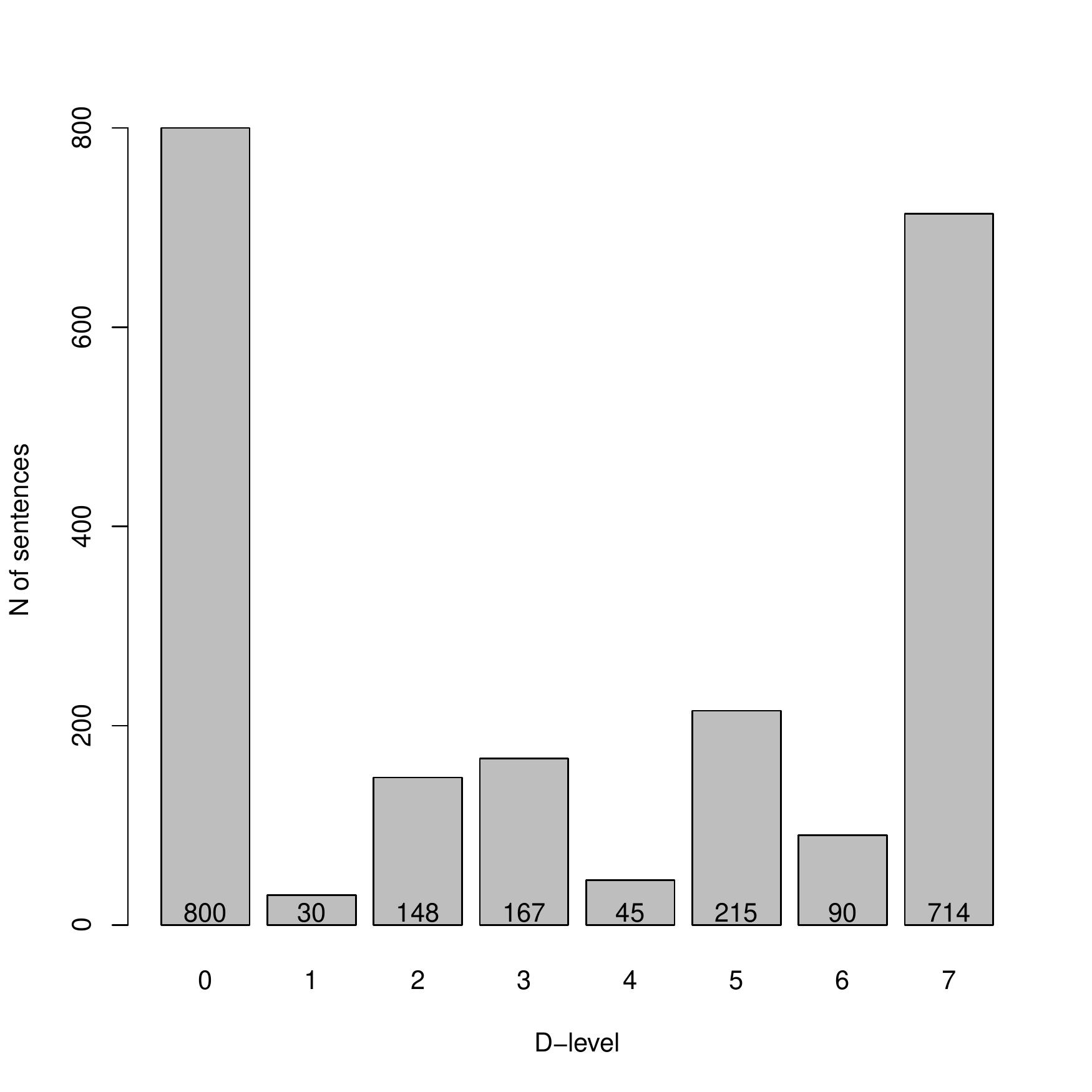}
\end{center}
\end{figure}

\subsection{D-level} 
The summary results for the D-level of 2209 WoW sentences are $M_{WoW}=3.45; SD_{WoW}=7.6$. Here, it is more interesting to look at the distribution of scores, which resembles a U-shape. In graph \ref{d-level-wow}, most sentences can be described either as level 0 or 7. There are 800 simple sentences and a roughly equivalent number of very complex sentences. This result is in line with the D-level distribution of scores on the game-{\it internal} ``quest'' texts in Thorne et al.~\cite{ThorneEtAl2012}. Our results need to be taken with a certain amount of reserve, however, due to the fact that upon manual inspection, several lists, incoherent productions, and errors from parsing and tagging were found. D-0 sentence are thus likely to be over-represented. 714 sentences were analyzed as complex, i.e. containing more than one structure from the level 1-6. Other sentences at the bottom of this distribution were characterized as sentences with subordinating conjunctions and non-finite clauses in adjunct position (215); sentences with relative clauses modifying verb object, finite clauses as verb objects and a few other types (167); sentences with conjoined constructions and co-ordinations (148) (see table \ref{D-levels} in Appendix for a detailed description of levels). Levels 1, 4 and 6 were rather infrequent, which could be attributed to how the categories were formed -- each is a rather narrow category (subject nominalizations, appositions, non-finite complements with own subjects, \ldots).


\section{Conclusion}
In this paper, we have presented a brief language-complexity description of the WoW-community texts, and compared those results to the general web corpus ukWaC. Overall, results show that the language production in the WoW corpus is slightly less complex, but not for all measures. Both corpora display the same amount of information packaged in nouns, but the WoW language can be described as more spoken-like, because its lexical density is lower in general. The lexical diversity as measured by TTR on samples reveals that the vocabulary size between texts in the WoW corpus differs to a larger extent than between texts in ukWaC, but on average, WoW texts reach a lower lexical diversity. The readability analysis showed that WoW texts form a homogeneous group with less dispersion than ukWaC texts. On the basis of readability score and average sentence length, the WoW texts can be described as intermediate or ``standard-style'' texts, appropriate for use with the 12--14 year-old students or older. The analysis of developmental levels, which enabled syntactic analysis of types of sentence embeddings, showed that for the most part, sentence production is either very simple or very complex. To sum up, our analysis shows very diverse results, reflecting the reality of the WoW-community language. Although the ukWaC scores mostly suggest a slightly more complex language, this is understandable since a greater variety of texts is collected in ukWaC than in the WoW corpus. Latter is a corpus of a fairly narrow-scope, technical language (cf. \cite{gee2007good}).
\paragraph{Further work} This study looked at game-external texts, however internal texts and fan fiction are two types of texts that could reveal a different picture than the one presented here.
Especially promising avenue for future research is syntactic analysis. A deeper analysis could allow us to examine the most salient constructions (see \ref{syntactic_complexity}). In this research, we chose a general web corpus as a point of comparison for WoW texts, yet other text types could be used, depending on the research aim. By looking for example at academic English, one could research how the complexity of WoW texts compares to the learners' language production in formal settings.

\bibliographystyle{unsrt}
\bibliography{call}

\section*{Appendix}
\begin{figure}[h]
\centering
\begin{minipage}{.47\linewidth}
  \centering
  \caption{Flesch's~\protect\cite{flesch} analysis of the readability of adult reading materials}\label{flesch_grades}
  \includegraphics[width=\textwidth]{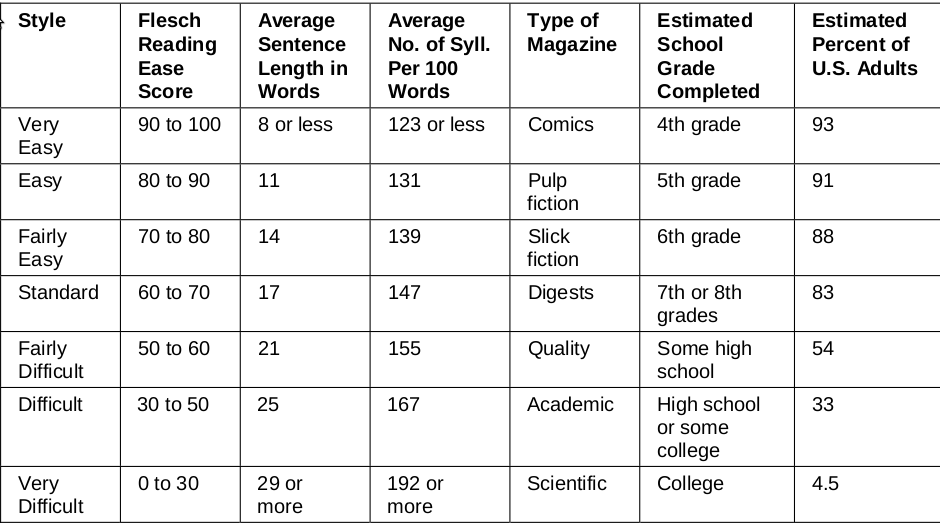}

\end{minipage}%
\hspace{0.5cm}
\begin{minipage}{.47\linewidth}
  \centering
  \caption{Developmental levels according to Covington et al.~\cite{Covington}. Table taken from Lu et al.~\cite{LuThorne2011}.}\label{D-levels}
  \includegraphics[width=\textwidth]{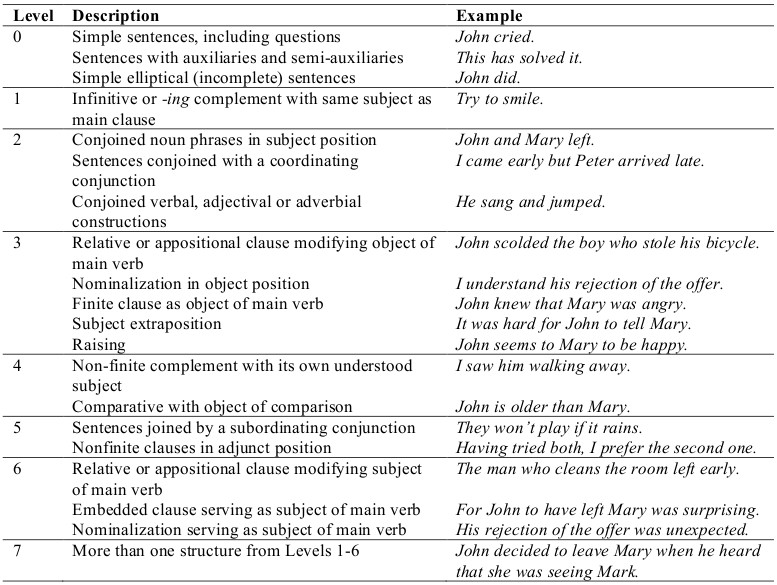}
\end{minipage}
\end{figure}

\end{document}